%% file: emnlp2col-t.tex
\def\TMACLIB{./}
\input{Macros.ttx}
\input \TMACLIB stdd.ttx 
\Refstda\Citenamedate 
\def\Underlinemark{\vrule height .7pt depth 0pt width 4.5pc}
\documentstyle[psfig,11pt,latex-acl]{article}
\input{syntax.tex}


\title{Valence Induction with a Head-Lexicalized PCFG}
\author{Glenn Carroll and Mats Rooth \\ IMS, Universit\"at Stuttgart
\\ {\tt \{glenn,mats\}@ims.uni-stuttgart.de}}

\begin{document}
\maketitle
\bibliographystyle{acl}

\section{Introduction}

Either directly or indirectly, the lexicon for a natural language
specifies {\em complementation frames} or {\em valences} for
open-class words such as verbs and nouns.  Constructing a lexicon of
complementation frames for large vocabularies constitutes a challenge
of scale, with the further complication that frame usage, like
vocabulary, varies with genre and undergoes ongoing innovation in a
living language. This paper addresses this problem by means of a
learning technique based on probabilistic lexicalized context free
grammars and the expectation-maximization (EM) algorithm.  Given a
hand-written grammar and a text corpus, frequencies of a head word
accompanied by a frame are estimated using the inside-outside
algorithm, and such frequencies are used to compute probability
parameters characterizing subcategorization.  The procedure can be
iterated for improved models. We show that the scheme is practical for
large vocabularies and accurate enough to capture differences in
usage, such as those characteristic of different domains.

\section{A grammar and formalism}

\begin{figure*}
\centerline{\psfig{figure=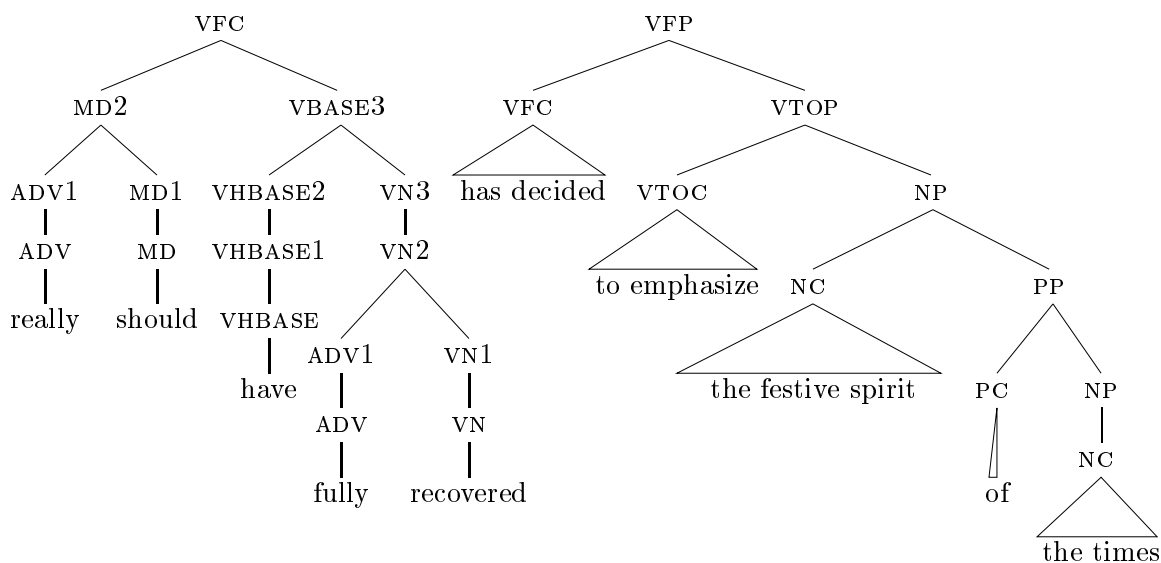}}
\caption{Illustrations of a finite verb 
chunk and complementation.}  
\label{max1}
\end{figure*}

The core of the grammar is an $\bar{X}$ grammar\Lspace \Lcitemark Jackendoff\Citebreak 1977\Rcitemark \Rspace{}
of phrases including noun phrases, prepositional phrases, and verbal
clusters.  A representative verbal structure is given on the left in
Figure \ref{max1}.  The symbol {\sc vfc} is read ``finite verb chunk'';
similarly we work with noun chunks ({\sc nc}), prepositional chunks
({\sc pc}), and so forth.  Our use of the chunk concept follows \LAcitemark Abney\Citebreak 1991\Citecomma
Abney\Citebreak 1995\RAcitemark{}. Categories are interpretable in
terms of a feature decomposition, but are treated as atomic in the
formalism.  We depart from a standard context-free formalism in that
heads are marked on the right hand sides of rules, using a prime (').

The grammar includes complementation rules for verbs, nouns, and
adjectives.  Complements are attached at a level above the chunk,
which we call the phrasal level.  For instance, the category {\sc vfp}
is expanded as a finite verb chunk {\sc vfc} and a sequence of
complements.  This is illustrated on the right in Figure 1, where the
{\sc vfc} headed by {\em decided} takes a {\sc vtop} complement, the
{\sc vtoc} headed by {\em emphasize} takes an {\sc np} complement, and
so forth.

\begin{figure*}
\centerline{\psfig{figure=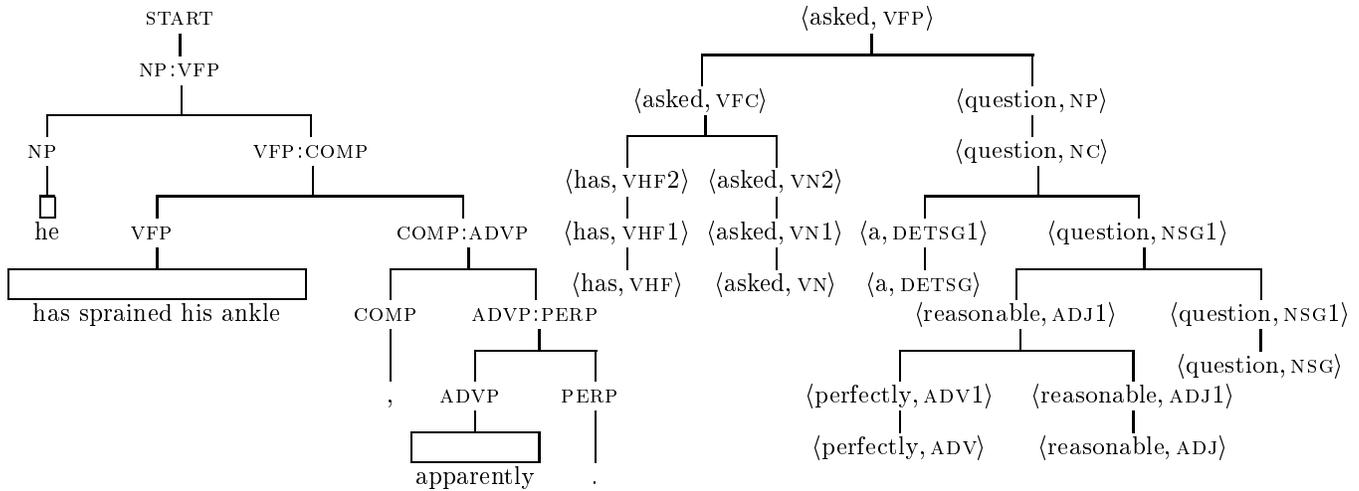,width=18cm}}
\caption{Left: finite-state structure;  Right: Lexicalization.} 
\label{TREE2}
\end{figure*}

Finally, the least standard part of the grammar is a large set of
{\em state} or {\em n-gram} rules which form a parse
without constructing a standard clause-level analysis.  Instead,
phrasal
categories are strung together with context-free rules modelling a
finite state machine, where the states are categories consisting
of an ordered pair of phrasal categories.  This results in
right-branching structures, as illustrated Figure 2.  Note that the
entire tree on the right in Figure 1 could be substituted for the
finite verb phrase {\sc vfp} in the tree on the left in Figure 
\ref{TREE2}.  The state rules 
allow almost all the sentences (about 97\%) in the corpus to be
parsed, at the price of not assigning linguistically realistic
higher-level structure.

We now define headed context-free grammars in the
sense employed here.

\vspace{0.5cm}
\noindent{\bf{Definition.}} A headed context free grammar is a tuple
$\tuple{N,T,W,{\cal L},{\cal R},s}$, where:
(i) $N$ and $T$ are disjoint sets, interpreted as the non-terminal
and terminal categories respectively.
(ii) $W$ is a set, interpreted as the set of words.
(iii) ${\cal L}$ is a relation between $W$ and $T$, indicating
the possible terminal categories (parts of speech) for a given 
word.
(iv) The set of headed productions ${\cal R}$ is a finite subset of 
$N \times  N^{*} \times (N \cup T) \times  N^{*}$, such that
each non-terminal occurs as the left hand side of some rule and 
each terminal occurs on the right hand side of some rule.
(v) $s \in N$, with the interpretation of a start symbol.

We typically use $\bar{n}$ as a variable for mother categories,
$n$  for head daughter categories, and $\alpha$ and
$\beta$ for the category sequences flanking the head on the right hand
side, so that $\tuple{\bar{n},\alpha,n,\beta}$ represents a rule.
$x$ is used as a variable for non-head categories.

A category $\bar{n}$ in $N$ is a {\em projection} of a category
$n$ in $N \cup T$ if there is some rule of the form
$\tuple{\bar{n},\alpha,n,\beta}$.  The set of {\em lexicalized
  nonterminals} ${\cal N} \subseteq W \times N$ is the composition of
${\cal L}$ with the transitive closure of the projection relation. We
have $\element{\pair{w}{n}}{{\cal N}}$ if the word $w$ can be the
lexical head of the nonterminal category $n$ (in a complete or
incomplete tree).

\section{Lexicalization and the probability model}
This section defines a parameterized family of probability
distributions over the trees licensed by a head-lexicalixed CFG.
The main ideas on the parameterization of a lexicalized context
free grammar which are employed here derive from 
\LAcitemark Charniak\Citebreak 1995\RAcitemark{}; see also the remarks on
lexicalization in 
\LAcitemark Charniak\Citebreak 1993\LIcitemark{}, section 8.4\RIcitemark \RAcitemark{}.

The head marking on rules
is used to project lexical items up a chain of categories.  In
the transitive verb phrase on the right in Figure 2, {\em question} is
projected to the {\sc np} level, and {\em asked} is projected to the
{\sc vfp} level.  In this tree, the non-terminal nodes are lexicalized
non-terminals, while the terminal nodes are members of ${\cal L}$.
The point of projecting head words is to make information which
probabilistically conditions rules and lexical choices available at
the relevant level.  At the top level in this example, the head {\em
  asked} is used to condition the choice of the phrase structure rule
{\sc vfp} $\rightarrow$ {\sc vfc}$'$ {\sc np} as well as the choice of
{\em question}, the head of the object.

We now define events which characterize choices of rules
and of lexical heads.

\noindent{\bf{Definition.}} Given a grammar 
$G=\tuple{N,T,W,{\cal L,R},s}$ with lexicalized non-terminals
${\cal N}$, the set of rule events $ER(G)$ is the set of tuples
$\tuple{w,\bar{n},\alpha,n,\beta}$ such that
$\pair{w}{\bar{n}}$ is an element of ${\cal N}$ and
$\tuple{\bar{n},\alpha,n,\beta}$ is an element of 
${\cal R}$.  The set of lexical choice events $EL(G)$ is the set of
tuples
$\tuple{w,\bar{n},x,v}$
such that
(i)  $\pair{w}{\bar{n}}$ and $\pair{v}{x}$ are elements of 
${\cal N}$;\footnote{
In the events, conditioning factors are ordered in the
way they are dropped off in the smoothing procedure described
below. In a lexical event $\tuple{w,\bar{n},x,v}$, the
choice of the word $v$ is conditioned on the parent lexical head
$w$, the parent category $\bar{n}$, and the child category
$x$.  In the first smoothing distribution, the first conditioning
factor, i.e. the parent head $w$, is dropped.
}
(ii) in some rule of the form $\tuple{{\bar{n}},\alpha,n,\beta}$,
$x$ is an element of one or both of the category sequences $\alpha$ and
$\beta$; and

By virtue of the length of tuples, $ER(G)$ and $EL(G)$ are disjoint,
and the union $E(G)$ can be formed without confusing lexical with rule
events.  

A head-lexicalized PCFG is represented as a function mapping
events to real numbers.

\noindent{\bf{Definition.}}  Let $G$ be a headed context free
grammar.  A head-lexicalized probabilistic context free grammar with
signature $G$ is a function $p$ with domain $E(G)$ and range $[0,1]$
satisfying the conditions:
(i) Fixing any lexicalized non-terminal $\pair{\bar{w}}{\bar{n}}$,
$\sum_{\alpha,n,\beta}p_{\bar{w},\bar{n},\alpha,n,\beta} = 1$;
(ii) Fixing any lexicalized non-terminal $\pair{\bar{w}}{\bar{n}}$ and
possible non-head daughter $x$, 
$\sum_{x,w}p_{\bar{w},\bar{n},x,w} = 1$.
Here the value of the function $p$ on a rule event is written as
$p_{\bar{w},{\bar{n}},\alpha,n,\beta}$, and on a lexical event as
$p_{{\bar{w}},{\bar{n}},x,w}$.

To assign probability weights to trees, we use a tree-licensing and
labelling interpretation of the grammar; a node in a tree analysis is
labeled with event corresponding to the rule used to expand the node,
and the list of lexical events for the non-head daughters of the node.
Where $\tau$ is a labeled tree licensed by G, we define $e(\tau): E(G)
\rightarrow \nat$ to be a function counting occurrences of events as
labels in $\tau$.  Algebraically, we think of $e(\tau)$ as a monomial
in the variables $E(G)$; the exponent of a given variable (or event)
$z$ is the number of occurrences of $z$ in $\tau$.  We denote the
evaluation of a polynomial or monomial $\phi$ in the variables $E(G)$
by subscripting: $\phi_{p}$ is the value of $\phi$ at the vector of
reals $p$.  Relative to a parameter setting $p$, $[e(\tau)]_{p}$ is
interpreted as the probabilistic weight of the labeled tree
$\tau$.\footnote{ As with ordinary PCFGs, depending on the parameters,
  the construction may or may not define a probability measure on the
  set of finite trees licensed by $G$.  For the general case, infinite
  trees can be included in the sample space.  This requires an
  extension in the definition of the measure but does not affect the
  probabilities of finite trees.  }

\begin{figure*}           
\centerline{\psfig{figure=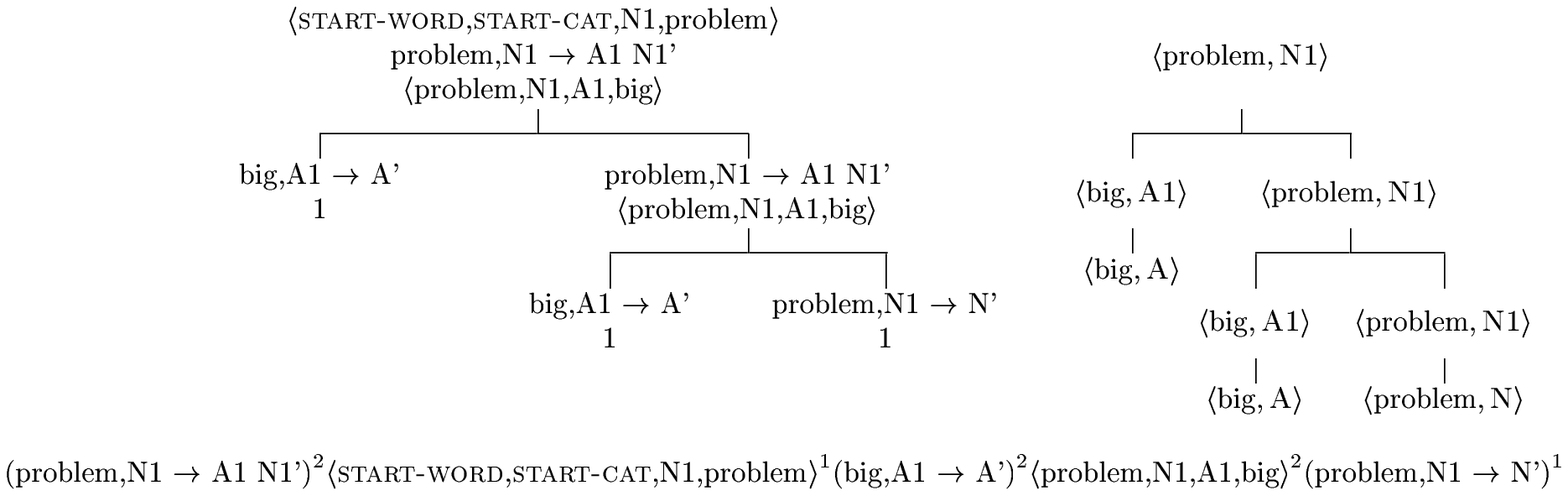,width=17cm}}
\caption{On the left, an event tree. On the right, the corresponding 
lexicalized tree. On the bottom, the event
monomial obtained as a symbolic product of the event labels.
The lexical choice event involving {\sc start-cat} chooses the
head of the sentence, in this case {\em problem.}}
\label{event-tree}
\end{figure*}

These notions are exemplified in Figure \ref{event-tree}, which is a
phrase structure tree for the N1 (read: N-bar) 
{\em big big problem} in a grammar where N1 is the sentence
category.  Each non-terminal is labeled with a phrase structure rule,
and with lexical choice events for non-head daughters. In this case,
the only non-head daughters are the two A1's headed with head {\em
  big}.  $\tuple{\mbox{problem,N1,A1,big}}$ is a lexical choice event
where {\em big} is selected as the head of an A1 with parent category
N1, and parent head {\em problem}. An event monomial corresponding to
the event tree is obtained as the symbolic product of the events
labeling the tree.

\section{Parameter Estimation}

Given a  grammar $G$, the inductive problem is to estimate
a head-lexicalized PCFG with signature $G$.  We work with the standard
method for estimating PCFGs, based on the Expectation-Maximization
framework
\Lcitemark Baum\Nameand Sell\Citebreak 1968\Citecomma
Dempster\Namecomma Laird\Nameandd Rubin\Citebreak 1977\Rcitemark \Rspace{}.

Above, we defined the event polynomial $e(\tau)$ for an
event tree $\tau$ licensed by $G$.  The event polynomial for a 
sentence $\sigma$ is the sum of the event polynomials 
for the event trees with yield $\sigma$.  Where corpus ${\cal C}$ is
a sequences of sentences, the corpus event polynomial $e({\cal C})$
is the (polynomial) product of the event polynomials for the sentences in 
${\cal C}$.  In these terms, maximum likelihood estimation selects
a parameter setting $p$ such that the value $[e({\cal C})]_{p}$
of the corpus polynomial is maximized; this corresponds to selecting
a parameter setting which maximizes the probability of the corpus.

The E step of the EM algorithm computes an expected event count
function which can be defined in terms of the corpus polynomial. In
the estimation of PCFGs using the inside-outside algorithm, event
counts are computed iteratively, sentence by sentence. The computation
uses a packed parse forest, a compact and-or graph representing
a set of trees and the sentence event polynomial, and which allows
efficient computation of expected event counts.  Somewhat more formally,
we use the Inside-outside algorithm
\Lcitemark Baker\Citebreak 1979\Rcitemark \Rspace{}.
to compute
\( E_p( z | \sigma ): E(G) \rightarrow {I\hspace{-0.7ex}R} \) 
where $z$ ranges over events in the join rule and lexical event
space $E(G)$, defined earlier.  $c(\sigma,p)(z)$ has
the probabilistic interpretation of the expected number of occurrences
of the event $z$ in the set of trees with yield $\sigma$.  

Given a parameter setting $p$, event counts are computed and summed
over the sentences in the corpus. In the algorithm of Baum and Sell,
new parameter values would be defined as relative frequencies of event
counts, i.e. maximum-likelihood estimation based on hidden data in the
EM framework.  We use instead a modified M step involving a smoothing
scheme in order to deal with the size of the parameter space and the
resulting problems that (i) counts are zero for the majority of
events, and (ii) the parameter space is too large to be represented
directly in computer memory.  Lexicalized rules are smoothed against
non-lexicalized rules in a standard back-off scheme
\Lcitemark Katz\Citebreak 1980\Rcitemark \Rspace{}.  The smoothed
probability is defined as a weighted sum of the maximum-likelihood
estimates for the lexicalized and unlexicalized rule probabilities.
The smoothing weight is allowed to vary through five discrete values
as a function of the frequency of the word-category pair.  The
parameters give greater weight to the lexicalized distribution when
enough data is present to justify it.  The smoothing parameters are
set using the EM algorithm on reserved data.

For the lexical choice distributions, an absolute discounting scheme
from \LAcitemark Ney\Namecomma Essen\Nameandd Kneser\Citebreak 1994\RAcitemark{}
is used, which is similar to Good-Turing, but
somewhat simpler to work with.

\section{The experiment}

We estimated a head-lexicalized PCFG from parts of the British
National Corpus\Lspace \Lcitemark BNC Consortium\Citebreak 1995\Rcitemark \Rspace{}, using the grammar described in
the first section and the estimation method of the previous section.
A bootstrapping method was used, in which first a non-lexicalized
probabilistic model was used to collect lexicalized event counts.  On
the next iteration, counts were estimated based on a lexicalized
weighting of parses, as described in the previous section.

Analyses were restricted to those consistent with the part of speech
tags specified in the BNC, which are produced with a tagger.
In each lexicalized iteration, event counts
were collected over a contiguous five million word segment of the
corpus. Parameters were re-computed in the way described above, and
the procedure was iterated on the next contiguous five-million word
segment.  Results from all iterations were pooled to form a single
model estimated from 50M words. Table \ref{wordselection} illustrates
lexical distributions in this model.

This training scheme allows the frame distributions for high-frequency
words a chance to converge on their true distributions, whereas a
single 50M word iteration would not.  The strategy derives from a
variant generalized EM algorithm presented in \LAcitemark Neal\Nameand Hinton\Citebreak 1998\RAcitemark{}.
In a nutshell, re-estimating the parameters during the course of a
single training iteration will still lead to convergence on a 
maximum-likelihood estimate, provided certain conditions are met.
Foremost among these is the requirement that no parameter setting can
be prematurely set to zero;  this is met by our smoothing strategy.
This is not to say that precisely the same strategy, pursued across
multiple iterations, would produce a maximum-likelihood estimate;  it
would not.  However, ``classical'' EM, requiring repeated iteration over
the entire training set, is both relatively inefficient and
infeasible given our present computational resources.

\input{table1}


\subsection*{Dictionary Evaluation}

The comparison to frames specified in a dictionary we use was
introduced by 
\LAcitemark Brent\Citebreak 1993\RAcitemark{} and subsequently used by
\LAcitemark Manning\Citebreak 1993\RAcitemark{}, \LAcitemark Ersan\Nameand Charniak\Citebreak 1995\RAcitemark{}
and
\LAcitemark Briscoe\Nameand Carroll\Citebreak 1996\RAcitemark{}.
The measure uses {\em precision} and {\em recall} to compare the set
of induced frames to those in the standard. Precision is the
percentage of frames that the system proposes that are correct (i.e.
in the standard).  Recall is the percentage of frames in the standard
that the system proposes. If the results are broken down into true
positives (TP), false positives (FP), true negatives (TN), and false
negatives (FN), precision is defined as \( TP / (TP + FP) \) and
recall is \( TP / (TP + FN) \).  To produce measurements from our
system, we must first reduce our distributions to set
membership. Brent proposed a stochastic filter for this reduction,
consisting of a set of per-frame probability cutoffs, which are
applied independently of the lexical head.  Although though the
independence assumption is certainly dubious, we have adopted this
method, without change, except for the introduction of a 
heuristic for finding the frame cutoffs.

The key property of cutoffs is that they control the tradeoff of
precision versus recall.  Raising the cutoff will generally produce a
higher precision, but lower recall, and contrariwise.  As we are
neutral about this tradeoff, we set the cutoffs at the crossover
point, where the difference in precision and recall changes sign.
This is not entirely deterministic, as the measures may cross more
than once; in that case, we optimize for the best precision.

\input{table2}   

For our dictionary, we used {\em The Oxford Advanced Learner's
  Dictionary}
\Lcitemark Hornby\Citebreak 1985\Rcitemark \Rspace{}, also used by Ersan/Charniak and Manning. We reduced our frame set
and the dictionary's to a common set, mapping some frames and
eliminating others.  For evaluation, we selected 200 verbs at random
from among those that occurred more than 500 times in the training
data; half were used to set the optimal cutoff parameters, and
precision and recall were measured with the remainder.

Table \ref{PRECISION} shows results broken down by frame.  The largest
source of error is the intransitive frame.  It is not hard to
understand why: our robust parsing architecture resolves unparsable
constructs as intransitives.  In addition to sentences where verbs are
not linked up with their complements because of interjections, complex
conjunctions or ellipses, this includes frames such as {\sc sbar} and
{\sc wh-}complements which are not included in the chunk/phrase
grammar.  While it would be possible in principle to extract these
from the present word collocation statistics, we plan instead to
pursue a solution involving extensions in the grammar.

A second major source of error is prepositional phrases.
The complementation model embodied in the PCFG does not distinguish
complements from adjuncts, and therefore adjunct prepositional phrases
are a source of false positives.  Thus the {\sc np pp} frame is
scored as a false positive for the verb {\em meet}, because the OALD
does not list the frame, although the combination appears often in the
corpus data.  While such frames lead to a loss of precision
in the dictionary evaluation, we do not necessarily consider them
a flaw in the information learned by the system, since
the argument/adjunct distinction is often tenuous, and adjuncts are in many
cases lexically conditioned.

Lastly, there are many false negatives for the particle frame and noun
plus particle.  This is mainly due to disagreements between BNC particle
tagging and particle markup in the OALD.

Despite these difficulties, the summary shown in table \ref{PRECISION}
shows results that are on the whole favorable.  In comparison with
other work with a comparable number of frames (Manning, Ersan/Charniak),
the system is well ahead on recall and well behind on precision.  If
one takes the sum of precision and recall to be the final performance
indicator, than we are slightly ahead: 1.54 vs. 1.44 for Ersan and
1.33 for Manning.  Briscoe and Carroll's work, with ten times as many target
frames, is so different that the numbers may be regarded as
incomparable.

Obviously, precision and recall measured against a standard relies on
the completeness and accuracy of that standard.  In checking false
positives, Ersan and Charniak found that the OALD was
incomplete enough to have a serious impact on precision.
Symmetrically, false negatives conflate deficiencies in
the corpus with poor learning efficiency.  It is impossible to say
based on table \ref{PRECISION} which of the systems is more efficient at
learning.  While our system shows the best recall, this could be
attributed to our having the best training data.  Charniak used 40M
words of training data, comparable to our 50M, but his data was
homogeneous, all taken from the Wall Street Journal.  As we will show
below, frame usage varies across genres, so the BNC, which includes
texts from a wide variety of sources, shows more varied frame usage
than the WSJ, and thus provides better data for frame acquisition.

\input{table3.tex}
\section{Cross entropy evaluation}

The information-theoretic notion of {\em cross entropy}
provides a detailed measure of the similarity of the acquired
probabilistic lexicon to the distribution of 
frames actually exhibited in the corpus (which we call the
empirical distribution).
 The cross entropy of the
estimated distribution $q$ with the empirical distribution $p$ obeys
the identity
\[
CE(p,q) =  H(p) + D(p \| q)  
\]
where $H$ is the usual entropy function and $D$ is the relative entropy,
or Kullback-Leibler distance. 
The entropy of a distribution over frames
can be conceptualized as the average number of
bits required to designate a frame in an ideal code based on
the given distribution.  In this context, entropy
measures the complexity of the observed frame distribution.
The relative entropy is the penalty paid in bits
when the frame is chosen according to the empirical
distribution $p$, but the code is derived from the model's
estimated distribution, $q$.
Relative entropy is always non-negative, and reaches zero only when
the two distributions are identical.  
Our goal, then, is to minimize the relative
entropy.  For more in-depth discussion 
of entropy measures, see
\LAcitemark Cover\Nameand Thomas\Citebreak 1991\RAcitemark{}, or any introductory 
information theory text.

For relative entropy to be finite, the estimated distribution $q$
must be non-zero whenever $p$ is.  However, some observed frames are not
present in the grammar, for one of two reasons.  Some well-known
frames such as {\sc sbar} require high-level constructs not 
available in the chunk/phrase grammar and 
unusual/unorthodox frames turn up in the data, e.g.
{\sc part pp pp}.  Since the model lacks these frames, smoothing
against the unlexicalized rules is insufficient.  Instead, for all the
estimated distributions, we smooth
against a Poisson distribution over categories, which assigns
non-zero probability to all frames, observed or not.  This allows us
to spell out the unknown frame using a known finite alphabet, the 
grammar categories, while
retaining a reasonable average length over frames.

\begin{table}
\begin{center}
\begin{tabular}{|r|r|l|c|c|}
\hline
\multicolumn{2}{|c|}{obs freq } & & \multicolumn{2}{|c|}{est freq}\\ 
\cline{1-2} \cline{4-5}
imag    & natsci   & frame & imag & natsci\\ \hline \hline
51     & 39 & {\sc np vtop  }       & 40.4      & 34.2 \\
21      & 43  & {\sc np     }   & 20.7          & 33.1 \\
13       & 6  & {\sc np np }    & 8.8           & 3.9 \\
6       & 1   & {\sc np pp}     & 3.2           & 4.7 \\ 
5       & 1   & {\sc np part}   & 1.7           & 1.0 \\ 
2      & 11   & {\sc pp   }     & 1.8           & 10.2 \\
1     & 0    & {\sc sbar}       & 0             & 0 \\ 
1    & 0  &  Intrans            & 9.3           & 7.6 \\
\hline \hline
2.130 & 1.913 & entropy & 2.476 & 2.423\\
\hline
\end{tabular}
\end{center}
\caption{True and estimated frame frequencies for {\em allow}.}
\label{ALLOW}
\end{table}

For our entropy measurements,
we selected three verbs, {\em allow}, {\em reach}, and {\em suffer}
and extracted about 200 occurrences of each from portions of the BNC
not used for training.  
Half of each sample was drawn from ``imaginative'' text
and the other half from the natural or applied sciences, as indicated
by BNC text mark-up.  The true frame for each verb 
occurrence was marked by a human judge\footnote{For this judgment,
the frame set was unrestricted, i.e. included frames not in the grammar.}.
The empirical distribution was taken as the
maximum-likelihood estimate from these frequencies.  Tables 4 
and 5 indicate the observed frequencies and the entropy of
the resulting distributions. 

Alongside the observed frequencies, we indicate a set of estimated
frequencies.  These were generated by taking the 50M word model described
above, parsing the test sentences, and extracting the estimated
frequencies.  The sum of estimated frequencies is generally less than the
observed frequencies due to tagging errors, parse failures, and frequency
assigned to frames not shown in the tables.  However,
an eyeball inspection of the tables shows that the parser does
a good job of reproducing the target distribution.  

\begin{table}
\begin{center}
\begin{tabular}{|r|r|l|c|c|}
\hline
\multicolumn{2}{|c|}{obs freq } & & \multicolumn{2}{|c|}{est freq}\\ 
\cline{1-2} \cline{4-5}
imag    & natsci   & frame & imag & natsci \\ \hline \hline
63     & 88  & {\sc np   }      & 50.1  & 74.5 \\
13  & 15  & {\sc np pp    }     & 5.9   & 10.9 \\
9    & 1   & {\sc part     }    & 5.9   & 0.8 \\
6    & 0   & {\sc part pp     } & 2.7   & 0 \\
5      & 3  & {\sc pp       }   & 6.7   & 3.4 \\
4       & 1  & Intrans          & 15.2  & 6.8 \\
2       & 0   & {\sc part np }  & 0.5   & 0 \\
1       & 0   & {\sc np part }  & 0     & 0.1 \\
\hline \hline
2.0 & 0.979  & entropy & 2.101 & 1.473 \\
\hline
\end{tabular}

\vspace{0.2cm}

\begin{tabular}{|r|r|l|c|c|}
\hline
\multicolumn{2}{|c|}{obs freq } & & \multicolumn{2}{|c|}{est freq}\\ 
\cline{1-2} \cline{4-5}
imag & natsci & frame & imag & natsci \\ \hline \hline
41      & 6  & Intrans & 34.9 & 13.4 \\
31       & 54   & {\sc pp     }         & 27.4 & 50.5 \\
21       & 36   & {\sc np     }         & 18.9 & 23.0   \\
4       & 1   & {\sc np vtop}           & 2.1 & 0.7 \\
3       & 4   & {\sc np pp }    & 0.9   & 5.2 \\
\hline \hline 
1.936 & 1.580  & entropy & 1.936 & 1.907 \\
\hline
\end{tabular}
\end{center}
\caption{True and estimated frame frequencies for {\em reach} (top)
and
{\em suffer} (bottom).}
\label{SUFFER}
\end{table}

One striking feature in the tables is the variation across
genre.  In particular, {\em suffer} used in the imaginative
genre shows a very different distribution than {\em suffer} in the
natural sciences.   A chi-squared test applied to each
pair indicates that the samples come from distinct distributions 
(confidence $> 95\%$).  

The column labeled ``50M lex'' in Table \ref{FIRST-ENTROPY} provides a
quantitative measure of the agreement between the 50M word combined
model and the empirical distributions for the three verbs in two
genres in the form of relative entropy.  The first column repeats the
entropy of the data distributions. For purposes of comparison, the
second column indicates the relative entropy of one data distribution
with the other data distribution filling the role of the estimated
distribution (i.e. $q$) in the discussion above.  The
relative entropy is lower when the estimated distribution is used for
$q$ than when the data distribution for the other genre is used for
$q$ in each case but one, where the figures are the same.
This suggests the combined model contains fairly good
overall distributions.


\input{table6.tex}

To numerically evaluate whether the system was able to learn the
distribution exhibited in a given collection of sentences, we tuned the
lexicon by parsing the test sentences for each genre separately with
the 50M word model, extracting the frequencies, and estimating the
distribution from these.  The results are the column 4 labeled ``50M
lexicalized extraction'' in \ref{SECOND-ENTROPY}. The following columns give the
same figures for freqency extraction with other models.  Extraction
with the large lexicalized model gives the best results, and gives
better relative entropy than the 50M lexicalilazed model itself (in
column 2). Notice that only the distributions estimated with the two
50M models are better than the  50M lexicalized model, though the
unlexicalized one is only marginally better.  In this sense, only the
50M lexicalized parser proves to be a  good enough parser for
genre tuning.  Notice that with this model, tuning in no case gives
worse relative entropy, and in five out of six cases give an
improvement. 

Notice also that relative entropy for the distributions obtained by
tuning with the 50M model are a good deal lower than the cross-genre
figures from Table \ref{FIRST-ENTROPY}.  This suggests that if we wanted 
to have a
good probabilistic lexicon for, say, the imaginative genre, we would
be better off using the automatic extraction procedure on data drawn
from that genre than using a
{\em perfect} parser (or a lexicographer) on data drawn from some 
other genre, such as the natural sciences. 
This provides a calibration of the accuracy of the
lexicalized parser's estimates, and conversely demonstrates that
words are not used in the same way in different genres.

\input{table7.tex}

\section{Optimal parses} 
\begin{figure*}
\centerline{\psfig{figure=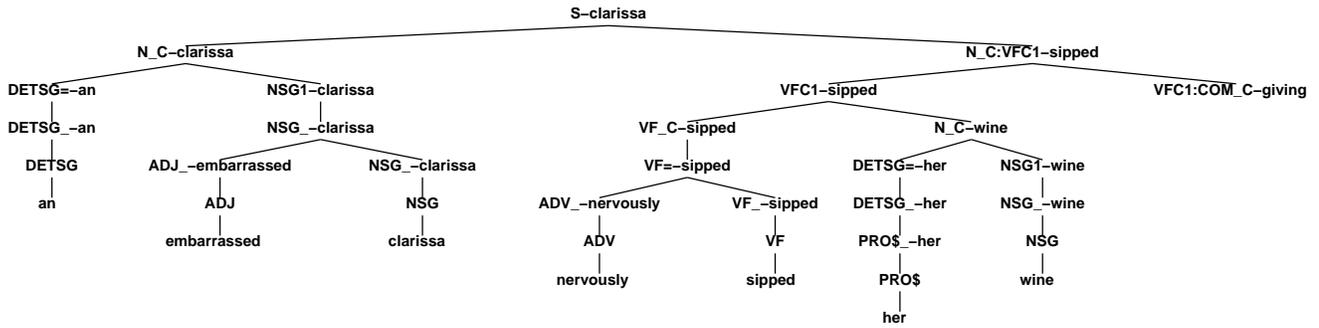,angle=270,width=18cm}}
\vspace{-1.6cm}
\caption{The first part of maximum probability parse.}  
\label{max2}
\end{figure*}

Although identifying a unique parse does not play a role in our
experiment, it is potentially useful for applications. A simple
criterion is to pick a parse with maximal probability; this is
identified in a parse forest by iterating from terminal nodes,
multiplying child probabilities and the local node weight at {\em
  and}-nodes (chart edges), and choosing a child with maximal 
  probability at {\em or}-nodes (chart constituents).  
  Figures \ref{max1} and \ref{max2} give examples of maximal
probability probability parses.

Other optimality criteria can be defined.  The structure on noun
chunks is often highly ambiguous, because of bracketing and part of
speech ambiguities among modifiers.  For many purposes, the internal
structure of an noun chunk is irrelevant; one just wants to identify
the chunk.  From this point of view, a probability estimate
which considers just one analysis might
underestimate the probability of a noun chunk.
In what we call a sum-max parse, probabilities are
summed within chunks by the inside algorithm.  Above the chunk
level, a highest-probability tree is computed, as described above.

\section{Notes on the implementation and parsing times}
Software is implemented in C++.
The parser used for the bootstrap phase is a vanilla CFG chart
parser, operating bottom-up with top-down predictive filtering.  
Chart entries are assigned probabilities using the unlexicalized PCFG,
and the lexicalized frequencies are found by carrying out a modified
inside-outside algorithm which simulates lexicalization of the chart.  

In the iterative training phase, an unlexicalized context-free skeleton is
found with the same parser.  We transform this into
its lexicalized form---categories become $\tuple{w,n}$ pairs and rules
acquire lexical heads---and carry out the standard inside-outside using the
more elaborate head-lexicalized PCFG model.  Average speed of the parser during
iterative training, including parsing, probability calculation, and 
recording observations, is 10.4 words per second on a Sun {\sc Sparc}-20.
The memory requirements for a model generated from a 5M word segment
are about 90Mbyte.  The upshot of all this is that we can train about
1M words per day on one machine, and a single 5M word iteration 
requires one machine work week.

\begin{figure*}
\centerline{\psfig{figure=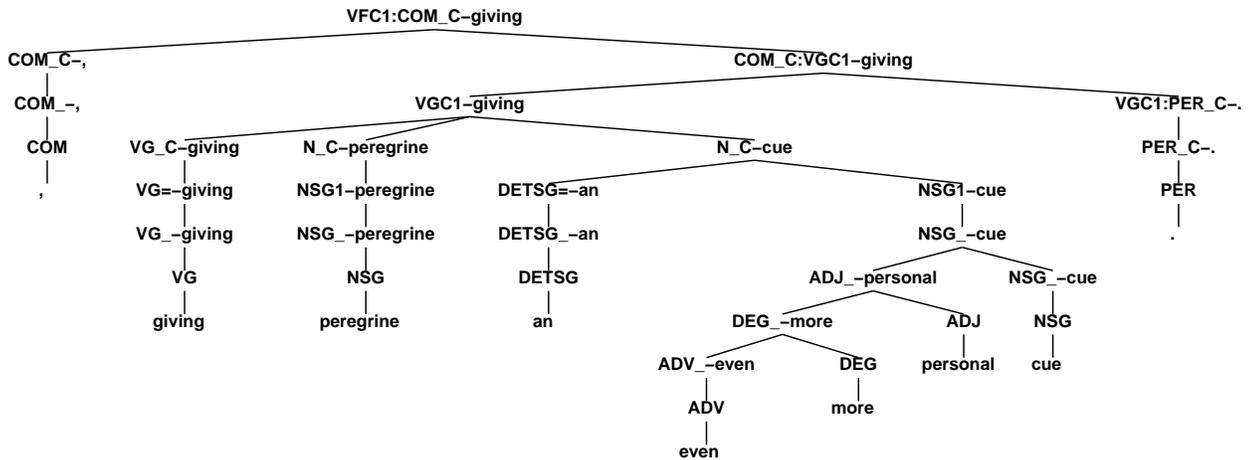,angle=270,width=18cm}}
\vspace{-1.8cm}
\caption{The second part.}  
\label{statetree}
\end{figure*}
\section{Discussion}
We believe the formalism and methodology described here have the
following advantages:

\begin{itemize}
\item  The grammar is under the control of the computational
linguist and is of a familiar kind, making it possible
to incorporate standard linguistic analyses, and making results
interpretable in terms of linguistic theory.  In contrast, approaches 
where context free rules are learned are likely to produce
structures which are uninterpretable in terms of linguistic
theory and practice.
\item Because of the context free framework, efficient parsing
algorthims (chart parsing) and probabilistic algorithms
(the inside-outside algorithm) can be applied.  With an efficient
implementation, this makes it possible to construct representations
of all the tree analyses for the sentences in corpora on the scale
of ten to a hundred million words, and to map such a corpus to
a probabilistic lexicon.
\item With the robustness introduced by the state model, almost
all sentences in the corpus can be parsed.
\item The model assigns probabilities to sentences and trees,
  which is useful for 
  applications independent of the lexicon-induction problem discussed
  here.
\item The word-selection model, which threads a word bigram model
through head relations in the syntactic tree, allows a large
body of word-word collocations to be learned from the corpus,
and put to use in weighting of competing analyses.
\item The valence information learned, rather than being
simply a set of subcategorization frames, is a probability
distribution which reflects the freqency of frames in a 
given training sample, and which can be plugged back into
the parser and used to analyze further text.
\end{itemize}

Some of these benefits are purchased at the cost of a lack of sophistication
in the grammar formalism, compared to constraint-based formalisms used
in contemporary computational linguistics.  This compromise is made in
order to make large-scale experiments achievable; our interest is in
conducting scientific experiments---observational and modeling 
experiments---with large bodies of
language use.  It is natural that this should require
incorporating approximations in computational models.
Notably, the compromises made in our approach are not so
severe that the grammatical analyses identified and the probability
parameters learned are out of touch with linguistic reality. This is
in contrast to the situation with other approaches using similar
mathematical methods, such as terminal-string n-gram
modeling.

\section{Conclusion}

We have presented a statistically-based method for valence induction,
based on the idea of automatic tuning of the probability parameters of
a grammar. On the standard precision/recall measures, our
system performs better on precision, worse on recall, and on the whole
somewhat better than other published systems.  We have provided a 
more precise evaluation via entropy measures, showing that the model
learns efficiently and builds accurate models of frame distributions.
The cross-domain entropy of the data frame distributions provides
numerical evidence that frame usage varies across domains, similar to
word usage.  This, in turn, suggests that automatic acquisition and
stochastic tuning are a must for large-scale NLP applications and
computational linguistic models.

 \section{Bibliography}
\message{REFERENCE LIST}

\bgroup\Resetstrings%
\def\Loccittest{}\def\Abbtest{ }\def\Capssmallcapstest{}\def\Edabbtest{}\def\Edcapsmallcapstest{}\def\Underlinetest{ }%
\def\NoArev{1000}\def\NoErev{0}\def\Acnt{1}\def\Ecnt{2}\def\acnt{0}\def\ecnt{0}%
\def\Ftest{ }\def\Ftrail{1}\def\Fstr{Abney\Citebreak 1991}%
\def\Atest{ }\def\Astr{Abney\Revcomma S\Initper }%
\def\Ttest{ }\def\Tstr{Parsing by Chunks}%
\def\Dtest{ }\def\Dstr{1991}%
\def\Btest{ }\def\Bstr{Views on Phrase Structure}%
\def\Etest{ }\def\Estr{D. Bouchard%
  \Eand K. Leffel}%
\def\Itest{ }\def\Istr{Kluwer Academic Publishers}%
\Refformat\egroup%

\bgroup\Resetstrings%
\def\Loccittest{}\def\Abbtest{ }\def\Capssmallcapstest{}\def\Edabbtest{}\def\Edcapsmallcapstest{}\def\Underlinetest{ }%
\def\NoArev{1000}\def\NoErev{0}\def\Acnt{1}\def\Ecnt{3}\def\acnt{0}\def\ecnt{0}%
\def\Ftest{ }\def\Ftrail{5}\def\Fstr{Abney\Citebreak 1995}%
\def\Atest{ }\def\Astr{Abney\Revcomma S\Initper }%
\def\Ttest{ }\def\Tstr{Chunks and dependencies:  Bringing processing evidence to bear on syntax}%
\def\Btest{ }\def\Bstr{Linguistics and Computation}%
\def\Etest{ }\def\Estr{Jennifer S. Cole%
  \Ecomma Georgia M. Green%
  \Eandd Jerry L. Morgan}%
\def\Itest{ }\def\Istr{CSLI Publications}%
\def\Dtest{ }\def\Dstr{1995}%
\def\Astr{\Underlinemark}%
\Refformat\egroup%

\bgroup\Resetstrings%
\def\Loccittest{}\def\Abbtest{ }\def\Capssmallcapstest{}\def\Edabbtest{}\def\Edcapsmallcapstest{}\def\Underlinetest{ }%
\def\NoArev{1000}\def\NoErev{0}\def\Acnt{0}\def\Ecnt{0}\def\acnt{0}\def\ecnt{0}%
\def\Ftest{ }\def\Ftrail{5}\def\Fstr{BNC Consortium\Citebreak 1995}%
\def\Ttest{ }\def\Tstr{The British National Corpus}%
\def\Otest{ }\def\Ostr{http://info.ox.ac.uk/bnc/}%
\def\Dtest{ }\def\Dstr{1995}%
\def\Ctest{ }\def\Cstr{Oxford University}%
\def\Itest{ }\def\Istr{BNC Consortium}%
\Refformat\egroup%

\bgroup\Resetstrings%
\def\Loccittest{}\def\Abbtest{ }\def\Capssmallcapstest{}\def\Edabbtest{}\def\Edcapsmallcapstest{}\def\Underlinetest{ }%
\def\NoArev{1000}\def\NoErev{0}\def\Acnt{1}\def\Ecnt{2}\def\acnt{0}\def\ecnt{0}%
\def\Ftest{ }\def\Ftrail{9}\def\Fstr{Baker\Citebreak 1979}%
\def\Atest{ }\def\Astr{Baker\Revcomma J\Initper \Initgap K\Initper }%
\def\Ttest{ }\def\Tstr{Trainable grammars for speech recognition}%
\def\Jtest{ }\def\Jstr{Proceedings of the Spring Conference of the Acoustical Society of America}%
\def\Dtest{ }\def\Dstr{1979}%
\def\Etest{ }\def\Estr{Jared J. Wolf%
  \Eand Dennis H. Klatt}%
\def\Ptest{ }\def\Pcnt{ }\def\Pstr{547--550}%
\def\Itest{ }\def\Istr{MIT}%
\def\Ctest{ }\def\Cstr{Cambridge, MA}%
\Refformat\egroup%

\bgroup\Resetstrings%
\def\Loccittest{}\def\Abbtest{ }\def\Capssmallcapstest{}\def\Edabbtest{}\def\Edcapsmallcapstest{}\def\Underlinetest{ }%
\def\NoArev{1000}\def\NoErev{0}\def\Acnt{2}\def\Ecnt{0}\def\acnt{0}\def\ecnt{0}%
\def\Ftest{ }\def\Ftrail{8}\def\Fstr{Baum\Nameand Sell\Citebreak 1968}%
\def\Atest{ }\def\Astr{Baum\Revcomma L\Initper \Initgap E\Initper %
  \Aand Sell\Revcomma G\Initper \Initgap R\Initper }%
\def\Ttest{ }\def\Tstr{Growth Transformations for Functions on Manifolds}%
\def\Jtest{ }\def\Jstr{Pacific Journal of Mathematics}%
\def\Dtest{ }\def\Dstr{1968}%
\def\Vtest{ }\def\Vstr{27}%
\def\Ntest{ }\def\Nstr{2}%
\Refformat\egroup%

\bgroup\Resetstrings%
\def\Loccittest{}\def\Abbtest{ }\def\Capssmallcapstest{}\def\Edabbtest{}\def\Edcapsmallcapstest{}\def\Underlinetest{ }%
\def\NoArev{1000}\def\NoErev{0}\def\Acnt{1}\def\Ecnt{0}\def\acnt{0}\def\ecnt{0}%
\def\Ftest{ }\def\Ftrail{3}\def\Fstr{Brent\Citebreak 1993}%
\def\Atest{ }\def\Astr{Brent\Revcomma M\Initper \Initgap R\Initper }%
\def\Ttest{ }\def\Tstr{From Grammar to Lexicon:  Unsupervised Learning of Lexical Syntax}%
\def\Jtest{ }\def\Jstr{Computational Linguistics}%
\def\Vtest{ }\def\Vstr{19}%
\def\Ntest{ }\def\Nstr{2}%
\def\Ptest{ }\def\Pcnt{ }\def\Pstr{243--262}%
\def\Itest{ }\def\Istr{Association for Computational Linguistics}%
\def\Dtest{ }\def\Dstr{1993}%
\Refformat\egroup%

\bgroup\Resetstrings%
\def\Loccittest{}\def\Abbtest{ }\def\Capssmallcapstest{}\def\Edabbtest{}\def\Edcapsmallcapstest{}\def\Underlinetest{ }%
\def\NoArev{1000}\def\NoErev{0}\def\Acnt{2}\def\Ecnt{0}\def\acnt{0}\def\ecnt{0}%
\def\Ftest{ }\def\Ftrail{6}\def\Fstr{Briscoe\Nameand Carroll\Citebreak 1996}%
\def\Atest{ }\def\Astr{Briscoe\Revcomma T\Initper %
  \Aand Carroll\Revcomma J\Initper }%
\def\Ttest{ }\def\Tstr{Automatic Extraction of Subcategorization from Corpora}%
\def\Dtest{ }\def\Dstr{1996}%
\def\Jtest{ }\def\Jstr{MS }%
\def\Otest{ }\def\Ostr{http://www.cl.cam.ac.uk/users/ejb/}%
\Refformat\egroup%

\bgroup\Resetstrings%
\def\Loccittest{}\def\Abbtest{ }\def\Capssmallcapstest{}\def\Edabbtest{}\def\Edcapsmallcapstest{}\def\Underlinetest{ }%
\def\NoArev{1000}\def\NoErev{0}\def\Acnt{1}\def\Ecnt{0}\def\acnt{0}\def\ecnt{0}%
\def\Ftest{ }\def\Ftrail{3}\def\Fstr{Charniak\Citebreak 1993}%
\def\Atest{ }\def\Astr{Charniak\Revcomma E\Initper }%
\def\Ttest{ }\def\Tstr{Statistical Language Learning}%
\def\Dtest{ }\def\Dstr{1993}%
\def\Itest{ }\def\Istr{MIT}%
\def\Ctest{ }\def\Cstr{Cambridge, MA}%
\Refformat\egroup%

\bgroup\Resetstrings%
\def\Loccittest{}\def\Abbtest{ }\def\Capssmallcapstest{}\def\Edabbtest{}\def\Edcapsmallcapstest{}\def\Underlinetest{ }%
\def\NoArev{1000}\def\NoErev{0}\def\Acnt{1}\def\Ecnt{0}\def\acnt{0}\def\ecnt{0}%
\def\Ftest{ }\def\Ftrail{5}\def\Fstr{Charniak\Citebreak 1995}%
\def\Atest{ }\def\Astr{Charniak\Revcomma E\Initper }%
\def\Ttest{ }\def\Tstr{Parsing with Context-free Grammars and Word Statistics}%
\def\Dtest{ }\def\Dstr{1995}%
\def\Rtest{ }\def\Rstr{Technical Report CS-95-28}%
\def\Itest{ }\def\Istr{Department of Computer Science, Brown University}%
\def\Astr{\Underlinemark}%
\Refformat\egroup%

\bgroup\Resetstrings%
\def\Loccittest{}\def\Abbtest{ }\def\Capssmallcapstest{}\def\Edabbtest{}\def\Edcapsmallcapstest{}\def\Underlinetest{ }%
\def\NoArev{1000}\def\NoErev{0}\def\Acnt{2}\def\Ecnt{0}\def\acnt{0}\def\ecnt{0}%
\def\Ftest{ }\def\Ftrail{1}\def\Fstr{Cover\Nameand Thomas\Citebreak 1991}%
\def\Atest{ }\def\Astr{Cover\Revcomma T\Initper \Initgap M\Initper %
  \Aand Thomas\Revcomma J\Initper \Initgap A\Initper }%
\def\Ttest{ }\def\Tstr{Elements of Information Theory}%
\def\Dtest{ }\def\Dstr{1991}%
\def\Itest{ }\def\Istr{John Wiley and Sons, Inc.}%
\def\Ctest{ }\def\Cstr{New York}%
\Refformat\egroup%

\bgroup\Resetstrings%
\def\Loccittest{}\def\Abbtest{ }\def\Capssmallcapstest{}\def\Edabbtest{}\def\Edcapsmallcapstest{}\def\Underlinetest{ }%
\def\NoArev{1000}\def\NoErev{0}\def\Acnt{3}\def\Ecnt{0}\def\acnt{0}\def\ecnt{0}%
\def\Ftest{ }\def\Ftrail{7}\def\Fstr{Dempster\Namecomma Laird\Nameandd Rubin\Citebreak 1977}%
\def\Atest{ }\def\Astr{Dempster\Revcomma A\Initper \Initgap P\Initper %
  \Acomma Laird\Revcomma N\Initper \Initgap M\Initper %
  \Aandd Rubin\Revcomma D\Initper \Initgap B\Initper }%
\def\Ttest{ }\def\Tstr{Maximum likelihood from incomplete data via the EM algorithm}%
\def\Jtest{ }\def\Jstr{Journal of the Royal Statistics Society}%
\def\Otest{ }\def\Ostr{Series B}%
\def\Dtest{ }\def\Dstr{1977}%
\def\Vtest{ }\def\Vstr{39}%
\def\Ptest{ }\def\Pcnt{ }\def\Pstr{1--38}%
\Refformat\egroup%

\bgroup\Resetstrings%
\def\Loccittest{}\def\Abbtest{ }\def\Capssmallcapstest{}\def\Edabbtest{}\def\Edcapsmallcapstest{}\def\Underlinetest{ }%
\def\NoArev{1000}\def\NoErev{0}\def\Acnt{2}\def\Ecnt{0}\def\acnt{0}\def\ecnt{0}%
\def\Ftest{ }\def\Ftrail{5}\def\Fstr{Ersan\Nameand Charniak\Citebreak 1995}%
\def\Atest{ }\def\Astr{Ersan\Revcomma M\Initper %
  \Aand Charniak\Revcomma E\Initper }%
\def\Ttest{ }\def\Tstr{A Statistical Syntactic Disambiguation Program and what it learns}%
\def\Dtest{ }\def\Dstr{1995}%
\def\Rtest{ }\def\Rstr{Brown CS Tech Report CS-95-29}%
\Refformat\egroup%

\bgroup\Resetstrings%
\def\Loccittest{}\def\Abbtest{ }\def\Capssmallcapstest{}\def\Edabbtest{}\def\Edcapsmallcapstest{}\def\Underlinetest{ }%
\def\NoArev{1000}\def\NoErev{0}\def\Acnt{1}\def\Ecnt{0}\def\acnt{0}\def\ecnt{0}%
\def\Ftest{ }\def\Ftrail{5}\def\Fstr{Hornby\Citebreak 1985}%
\def\Atest{ }\def\Astr{Hornby\Revcomma A\Initper \Initgap S\Initper }%
\def\Ttest{ }\def\Tstr{Oxford Advanced Learner's Dictionary of Current English}%
\def\Ctest{ }\def\Cstr{Oxford}%
\def\Itest{ }\def\Istr{Oxford University Press}%
\def\Dtest{ }\def\Dstr{1985}%
\def\Otest{ }\def\Ostr{4th Ed.}%
\Refformat\egroup%

\bgroup\Resetstrings%
\def\Loccittest{}\def\Abbtest{ }\def\Capssmallcapstest{}\def\Edabbtest{}\def\Edcapsmallcapstest{}\def\Underlinetest{ }%
\def\NoArev{1000}\def\NoErev{0}\def\Acnt{1}\def\Ecnt{0}\def\acnt{0}\def\ecnt{0}%
\def\Ftest{ }\def\Ftrail{7}\def\Fstr{Jackendoff\Citebreak 1977}%
\def\Atest{ }\def\Astr{Jackendoff\Revcomma R\Initper }%
\def\Ttest{ }\def\Tstr{$\overline{X}$ syntax: A study in phrase structure.}%
\def\Itest{ }\def\Istr{MIT Press}%
\def\Dtest{ }\def\Dstr{1977}%
\def\Ctest{ }\def\Cstr{Cambridge, MA}%
\Refformat\egroup%

\bgroup\Resetstrings%
\def\Loccittest{}\def\Abbtest{ }\def\Capssmallcapstest{}\def\Edabbtest{}\def\Edcapsmallcapstest{}\def\Underlinetest{ }%
\def\NoArev{1000}\def\NoErev{0}\def\Acnt{1}\def\Ecnt{0}\def\acnt{0}\def\ecnt{0}%
\def\Ftest{ }\def\Ftrail{0}\def\Fstr{Katz\Citebreak 1980}%
\def\Atest{ }\def\Astr{Katz\Revcomma S\Initper \Initgap M\Initper }%
\def\Dtest{ }\def\Dstr{1980}%
\def\Ttest{ }\def\Tstr{Estimation of probabilities from sparse data for the language model component of a speech recognizer}%
\def\Jtest{ }\def\Jstr{IEEE Transactions on Acoustics, Speech and Signal Processing}%
\def\Vtest{ }\def\Vstr{35}%
\def\Ptest{ }\def\Pcnt{ }\def\Pstr{400--401}%
\Refformat\egroup%

\bgroup\Resetstrings%
\def\Loccittest{}\def\Abbtest{ }\def\Capssmallcapstest{}\def\Edabbtest{}\def\Edcapsmallcapstest{}\def\Underlinetest{ }%
\def\NoArev{1000}\def\NoErev{0}\def\Acnt{1}\def\Ecnt{0}\def\acnt{0}\def\ecnt{0}%
\def\Ftest{ }\def\Ftrail{3}\def\Fstr{Manning\Citebreak 1993}%
\def\Atest{ }\def\Astr{Manning\Revcomma C\Initper }%
\def\Ttest{ }\def\Tstr{Automatic acquisition of a large subcategorization dictionary from corpora}%
\def\Dtest{ }\def\Dstr{1993}%
\def\Jtest{ }\def\Jstr{Proceedings of the 31st Annual Meeting of the ACL}%
\def\Itest{ }\def\Istr{Association for Computational Linguistics}%
\def\Ptest{ }\def\Pcnt{ }\def\Pstr{235--242}%
\Refformat\egroup%

\bgroup\Resetstrings%
\def\Loccittest{}\def\Abbtest{ }\def\Capssmallcapstest{}\def\Edabbtest{}\def\Edcapsmallcapstest{}\def\Underlinetest{ }%
\def\NoArev{1000}\def\NoErev{0}\def\Acnt{2}\def\Ecnt{1}\def\acnt{0}\def\ecnt{0}%
\def\Ftest{ }\def\Ftrail{8}\def\Fstr{Neal\Nameand Hinton\Citebreak 1998}%
\def\Atest{ }\def\Astr{Neal\Revcomma R\Initper \Initgap M\Initper %
  \Aand Hinton\Revcomma G\Initper \Initgap E\Initper }%
\def\Ttest{ }\def\Tstr{A New View of the EM Algorithm that Justifies Incremental and Other Variants}%
\def\Btest{ }\def\Bstr{Learning in Graphical Models}%
\def\Etest{ }\def\Estr{Michael I. Jordan}%
\def\Dtest{ }\def\Dstr{1998}%
\def\Itest{ }\def\Istr{Kluwer Academic Press}%
\Refformat\egroup%

\bgroup\Resetstrings%
\def\Loccittest{}\def\Abbtest{ }\def\Capssmallcapstest{}\def\Edabbtest{}\def\Edcapsmallcapstest{}\def\Underlinetest{ }%
\def\NoArev{1000}\def\NoErev{0}\def\Acnt{3}\def\Ecnt{0}\def\acnt{0}\def\ecnt{0}%
\def\Ftest{ }\def\Ftrail{4}\def\Fstr{Ney\Namecomma Essen\Nameandd Kneser\Citebreak 1994}%
\def\Atest{ }\def\Astr{Ney\Revcomma H\Initper %
  \Acomma Essen\Revcomma U\Initper %
  \Aandd Kneser\Revcomma R\Initper }%
\def\Ttest{ }\def\Tstr{On structuring probabilistic dependences in stochastic language modelling}%
\def\Jtest{ }\def\Jstr{Computer Speech and Language}%
\def\Dtest{ }\def\Dstr{1994}%
\def\Vtest{ }\def\Vstr{8}%
\def\Ptest{ }\def\Pcnt{ }\def\Pstr{1--38}%
\def\Itest{ }\def\Istr{Academic Press Limited}%
\Refformat\egroup%

\end{document}

%% file: syntax.tex
\newcounter{exampleno}
{\refstepcounter{exampleno} 
 \begin{list}{}{\setlength{\leftmargin}{0.45in}
                \setlength{\topsep}{0.1in}
                \setlength{\partopsep}{0.0in}
                \setlength{\itemsep}{0.0in}
                \setlength{\labelwidth}{0.3in}
                \setlength{\parsep}{0.0in}
                \setlength{\labelsep}{0.1in}}
 \item[(\theexampleno)\hfill]}%
{\end{list}}
{\refstepcounter{exampleno} 
 \begin{list}{}{\setlength{\leftmargin}{0.0in}
                \setlength{\topsep}{0.1in}
                \setlength{\partopsep}{0.0in}
                \setlength{\itemsep}{0.0in}
                \setlength{\labelwidth}{0.0in}
                \setlength{\parsep}{0.0in}
                \setlength{\labelsep}{0.1in}}
 \item[(\theexampleno)\hfill]}%
{\end{list}}
{\refstepcounter{exampleno} 
 \begin{list}{}{\setlength{\leftmargin}{0.55in}
                \setlength{\topsep}{0.1in}
                \setlength{\partopsep}{0.0in}
                \setlength{\itemsep}{0.0in}
                \setlength{\labelwidth}{0.4in}
                \setlength{\parsep}{0.0in}
                \setlength{\labelsep}{0.1in}}}%
{\end{list}}

\newcommand{\ignore}[1]{}

\newcommand{\nat}[0]{I\hspace{-0.7ex}N}

\newcommand{\pair}[2]{\left\langle#1,#2\right\rangle}

\newcommand{\tuple}[1]{\langle#1\rangle}

\newcommand{\element}[2]{#1 \epsilon #2 }



%% file: table1.tex
\begin{table}
\begin{center}
\hspace{1cm}{\begin{tabular}{|ll|ll|ll|}
\multicolumn{2}{c}{$p_{\mbox{\sc np},satisfactory,\mbox{\sc adjp},w}$}
            & \multicolumn{2}{c}{$p_{\mbox{\sc vfp},address,\mbox{\sc np},w}$} \\ \hline
 adverb & prob & noun & prob \\ \hline
  entirely&0.17 & question&0.086 \\
 highly&0.11  & issue&0.086 \\
  most&0.09& themselves&0.059 \\
 very&0.075& issues&0.031 \\
  quite&0.055& structure&0.031 \\
 wholly&0.032& argument&0.014 \\
 uncommonly&0.0037& questions&0.0043 \\
  especially&0.0037& electorate&0.0043\\ 
\multicolumn{2}{|c|}{\ldots}
            & \multicolumn{2}{c|}{\ldots} \\ \hline
\end{tabular}}
\end{center}
\caption{On the left: the eight largest
parameters in the lexical choice distribution
describing modifying
adjectives selected by {\em satisfactory}.  
On the right: parallel information for the distribution 
describing heads of
objects of the verb {\em address}.}
\label{wordselection}
\end{table}

%% file: table2.tex
\begin{table}
\begin{center}
\begin{tabular}{|l|c|c|c|c|c|c|}
\hline
         & cutoff & TP  &  FP  &  FN  &  prec  &  recl   \\ \hline
 Intrans  & 0.15 &  20  &  24  &  12  &  0.6471  &  0.7857 \\ \hline

 {\sc np }  & 0.021 &  3  &  5  &  1  &  0.9479  &  0.9891 \\ \hline

 {\sc adj }  & 0.079  &  92  &  0  &  6  &  1  &  0.25 \\ \hline

 {\sc pp }  & 0.045  &  27  &  15  &  6  &  0.7761  &  0.8966 \\ \hline

 {\sc part }  &  0.027  &  60  &  5  &  14  &  0.8077  &  0.6 \\ \hline

 {\sc vtop }  &  0.079  &  83  &  1  &  7  &  0.9  &  0.5625 \\ \hline

 {\sc np pp }  &  0.040  &  26  &  11  &  10  &  0.8281  &  0.8413 \\ \hline

 {\sc np part }  &  0.0099  &  68  &  6  &  12  &  0.7  &  0.5385 \\ \hline

 {\sc np np }  &  0.036  &  81  &  6  &  8  &  0.4545  &  0.3846 \\ \hline

 {\sc np vtop }  &  0.018  &  84  &  1  &  6  &  0.9  &  0.6 \\ \hline

 {\sc ving }  &  0.019  &  86  &  3  &  6  &  0.625  &  0.4545 \\ \hline

 {\sc np ving }  &  0.017  &  93  &  3  &  2  &  0.4  &  0.5 \\ \hline

 {\sc np vinf }  &  0.019  &  99  &  1  &  0  &  0  &  -- \\ \hline

 {\sc np adj }  &  0.016   &  85  &  1  &  12  &  0.6667  &  0.1429 \\ \hline

 {\sc pp vtop }  &  0.014  &  97  &  1  &  1  &  0.5  &  0.5 \\ \hline \hline

         & & 310  &  83  &  103  &  0.7888  &  0.7506 \\ \hline
\end{tabular} 
\end{center}
\label{PRECISION}
\caption{Precision/recall broken down by frame.}
\end{table}

%% file: table3.tex
\begin{table}
\begin{center}
\begin{tabular}{|l|c|c|c|}
\hline
 & precision\% & recall \% & no. of frames      \\
\hline
lex PCFG        & 79    &       75      & 15    \\ \hline
Briscoe         & 66    &       36      & 159   \\ \hline
Charniak        & 92    &       52      & 16    \\ \hline
Manning         & 90    &       43      & 19*   \\ \hline
\hline
\end{tabular}
\end{center}
\label{PR-COMPARISON}
\caption{Type precision/recall comparison.  Some of Manning's frames
are parameterized for a preposition.}
\end{table}

%% file: table6.tex
\begin{table}
\begin{center}
\begin{tabular}{|r|c|c|c|c|}
\multicolumn{2}{c}{} & \multicolumn{3}{c}{ $D( p \| q )$ for various $q$} \\
\cline{3-5}
\multicolumn{2}{c|}{} & other & 50M & 50M \\
\multicolumn{1}{c}{$head, genre$} & $H(p)$ & genre & lex & unlex  \\ \hline
imag &2.06&   0.50&0.40&3.13  \\
\raisebox{1.5ex}[0cm][0cm]{allow} natsci& 1.78&   0.49&0.42&2.27 \\
imag & 1.99 &  0.91&0.35&1.07  \\
\raisebox{1.5ex}[0cm][0cm]{reach} natsci& 0.90&   0.37&0.37&1.36 \\
imag & 1.86 &  0.87&0.24&0.70 \\
\raisebox{1.5ex}[0cm][0cm]{suffer} natsci& 1.51&   0.59&0.37&1.19 \\ \hline
 mean & 1.68&   0.62&0.36&1.62 \\
\end{tabular}
\end{center}

\caption{Frame relative entropy for three verbs in two genres. 
The first column names the lexical head and genre, and the second the
entropy ($H$) of the empirical distribution over frames, $p$.  By empirical
distribution we mean the relative frequencies from 
examples scored by a human judge.
Columns three through five give the relative entropy $D( p \| q)$ for
various related distributions.  In column three, $q$
is the empirical 
frame distribution for the same head, but with the complementary genre.
In column four $q$
is the (genre-independent) distribution derived from the 50M word 
lexicalized model.
Column five uses the unlexicalized
frame distribution derived from the 50M model, 
i.e. a distribution insensitive to the head verb.
Lower relative entropy is better.
}
\label{FIRST-ENTROPY}
\end{table}


%% file: table7.tex
\begin{table}
\begin{center}
\begin{tabular}{|r|c|c|c|c|c|}
\multicolumn{1}{c}{} & \multicolumn{5}{c}{ $D(p \| q)$ } \\
\cline{2-6}
\multicolumn{1}{c|}{} & 50M & 50M & 5M & 50M & 5M \\
\multicolumn{1}{c|}{} & lex & lex & lex & unl. & unl.\\
\multicolumn{1}{c|}{$head, genre$} & mod  & extr &  extr &  extr &  extr\\ \hline

imag & 0.40 & 0.32 & 1.32 & 0.47 & 1.32 \\
\raisebox{1.5ex}[0cm][0cm]{allow} natsci & 0.42 & 0.28 & 0.28 & 0.52 & 0.86 \\
imag & 0.35 & 0.35 & 0.63 & 0.32 & 0.63 \\
\raisebox{1.5ex}[0cm][0cm]{reach} natsci & 0.37 & 0.19 & 0.34 & 0.28 & 0.34 \\
imag & 0.24 & 0.11 & 0.38 & 0.12 & 0.38 \\
\raisebox{1.5ex}[0cm][0cm]{suffer} natsci & 0.37 & 0.20 & 0.88 & 0.34 & 0.88 \\ \hline
mean & 0.36 & 0.24 & 0.64 & 0.34 & 0.74 \\ 
\end{tabular}
\end{center}
\caption{Relative entropy of distributions estimated by parsing
the test sentences with various models, and 
using the Inside-outside algorithm to produce estimated 
distributions $q$.
The first column names empirical distributions $p$.
The second column repeats relative entropy for the 50M lexicalized model
from the previous table.  The third gives relative entropy 
where $q$ is obtained by
parsing and estimating frequencies in the test sentences 
with the 50M lexicalized model.
The following columns give the corresponding figures for a $q$
obtained by following the same procedure with a 5M word lexicalized model,
a 50M word unlexicalized model, and a 5M word unlexicalized model.
}
\label{SECOND-ENTROPY}
\end{table}
